\documentclass[lettersize,journal]{IEEEtran}
\usepackage{amsmath,amsfonts}
\usepackage{algorithmic}
\usepackage{algorithm}
\usepackage{array}
\usepackage[caption=false,font=normalsize,labelfont=sf,textfont=sf]{subfig}
\usepackage{textcomp}
\usepackage{stfloats}
\usepackage{url}
\usepackage{verbatim}
\usepackage{graphicx}
\usepackage{cite}
\usepackage{authblk}
\usepackage{rotfloat}

\begin{document}

\title{Explainable AI in Genomics: Transcription Factor Binding Site Prediction with Mixture of Experts}

\author{Aakash Tripathi, \IEEEmembership{Member,~IEEE}, Ian E. Nielsen, \IEEEmembership{Member,~IEEE}, Muhammad Umer, \IEEEmembership{Member,~IEEE}, Ravi P. Ramachandran,~\IEEEmembership{Senior Member,~IEEE} and Ghulam Rasool, \IEEEmembership{Senior Member,~IEEE}

\thanks{A. Tripathi and G. Rasool are with Machine Learning, Moffitt Cancer Center, 12902 USF Magnolia Drive, Tampa, FL, 33612, USA, (email: Aakash.Tripathi@moffitt.org, Ghulam.Rasool@moffitt.org).}
\thanks{I. E. Nielsen and R. P. Ramachandran are with the Department of Electrical and Computer Engineering, Rowan University, Glassboro, NJ, 08028, (e-mail: nielseni6@rowan.edu, ravi@rowan.edu).}
\thanks{M. Umer is with Electrical Engineering, Fairleigh Dickinson University, 1000 River Road, Teaneck, NJ, 07666, USA, (email: m.umer@fdu.edu)}
\thanks{This work was partly supported by NSF grants 2234468 and 2234836, and NIH grant P30-CA076292 to A. Tripathi and G. Rasool. I. E. Nielsen is supported by the US Department of Education through a Graduate Assistance in Areas of National Need (GAANN) program Award Number P200A180055.}}

\markboth{Preprint}%
{Shell \MakeLowercase{\textit{et al.}}: A Sample Article Using IEEEtran.cls for IEEE Journals}


\maketitle

\begin{abstract}
Transcription Factor Binding Site (TFBS) prediction is crucial for understanding gene regulation and various biological processes. This study introduces a novel Mixture of Experts (MoE) approach for TFBS prediction, integrating multiple pre-trained Convolutional Neural Network (CNN) models, each specializing in different TFBS patterns. We evaluate the performance of our MoE model against individual expert models on both in-distribution and out-of-distribution (OOD) datasets, using six randomly selected transcription factors (TFs) for OOD testing. Our results demonstrate that the MoE model achieves competitive or superior performance across diverse TF binding sites, particularly excelling in OOD scenarios. The Analysis of Variance (ANOVA) statistical test confirms the significance of these performance differences. Additionally, we introduce ShiftSmooth, a novel attribution mapping technique that provides more robust model interpretability by considering small shifts in input sequences. Through comprehensive explainability analysis, we show that ShiftSmooth offers superior attribution for motif discovery and localization compared to traditional Vanilla Gradient methods. Our work presents an efficient, generalizable, and interpretable solution for TFBS prediction, potentially enabling new discoveries in genome biology and advancing our understanding of transcriptional regulation.
\end{abstract}


\section{Introduction}
\IEEEPARstart{D}{eoxyribonucleic acid (DNA)} serves as the blueprint for the development and functionality of the human body. The information encoded in DNA is converted into functional molecules through the process of transcription, where DNA is transcribed into messenger RNA (mRNA) \cite{dreyfuss2002messenger}. Transcription is regulated by proteins called transcription factors that bind to specific DNA segments known as transcription factor binding sites (TFBS). TFBS can be located in gene promoters that initiate transcription or in cis-regulatory elements that modulate the transcription of nearby genes 
\cite{stormo2013modeling,    inukai2017transcription}.
These binding sites contain short, recurring DNA sequence motifs that determine the function of the associated DNA regions. Identifying TFBS motifs is essential for understanding fundamental biological processes, including transcriptional regulation, mRNA splicing, and the formation of protein complexes  
\cite{bailey2008discovering, vijayvargiya2013regulatory}.
Furthermore, motif identification facilitates the study of genetic diseases, gene evolution, and other key biological mechanisms \cite{prosperi2021fast}. Consequently, developing accurate and efficient machine-learning methods for TFBS motif identification is crucial for advancing our understanding of the genome.

Various machine learning approaches have been proposed for TFBS prediction \cite{zheng2021deep, zhang2021locating, cazares2023maxatac}. One simple yet powerful method is DeepBIND \cite{deepBIND}, which employs a convolutional neural network (CNN) architecture. Although DeepBIND achieves good results, CNNs are not optimally suited for processing sequential data like DNA. Later approaches, such as DeepRAM \cite{deepRAM}, have incorporated recurrent neural networks (RNNs) and long short-term memory (LSTM) architectures \cite{tavakoli2019modeling, singh2022splice, dasari2022explainable},
which are better adapted for sequential data. Recently, transformer-based architectures like DNABERT \cite{dnabert}, which utilize self-attention mechanisms, have achieved state-of-the-art performance on various genomic tasks.

Our preliminary experiments confirm that the DeepBind architecture provides strong baseline performance for TFBS prediction, making it an ideal building block for our proposed mixture of experts (MoE) approach where each expert utilizes this architecture. In our testing, we used a pre-trained DNABERT model to generate embeddings for 6-mer DNA sequences of GATA3 (GATA Binding Protein 3, a transcription factor crucial in T-cell development), processed by a transformer encoder (1 layer, 4 attention heads, 2048 feedforward dimension, 0.37 dropout rate). This DNABERT-based model achieved a test Area Under the Curve (AUC) of $0.7032$ for TFBS prediction. The AUC is calculated from the Receiver Operating Characteristic (ROC) curve, which plots the true positive rate against the false positive rate at various threshold settings. DeepBIND embeddings produced better results, achieving an AUC of $0.89$, likely due to the convolutional layers effectively capturing local sequence patterns. DeepBIND also offers advantages in training speed and interpretability, though it may have limitations in generalizing to diverse TFBS patterns.

To address this limitation of generalization, we propose a mixture of experts (MoE) approach that efficiently integrates multiple pre-trained CNN models, each specializing in different TFBS patterns. While RNN and LSTM architectures are well-suited for sequential data, we chose to use CNNs in our MoE approach due to their efficiency in capturing local motif patterns and their computational advantages in parallel processing, which are particularly beneficial for the large-scale genomic data involved in TFBS prediction. Our MoE model retains the benefits of CNNs for TFBS prediction while enhancing generalizability to unseen data. Additionally, we address the issue of explainability \cite{nielsen2022robust} through attribution mapping 
\cite{nielsen2023evalattai, ancona2019gradient} 
to gain user trust in the decision of the model. A novel attribution mapping technique, ShiftSmooth, is introduced that provides more robust model interpretability by considering small shifts in the input data. The paper presents several novel contributions:
\begin{enumerate}
    \item It introduces a new Mixture of Experts (MoE) model for TFBS prediction integrating multiple pre-trained CNN models for enhanced generalization. 
    \item It provides a comprehensive performance evaluation on in-distribution (same statistical distribution as the training samples) and out-of-distribution (referred to as OOD and whose statistical distribution is different from the training samples) data. This includes a comparison of all models by a statistical analysis using Analysis of Variance (ANOVA) \cite{anova} that exemplifies the superiority of the MoE approach.
    \item It analyzes both baseline and MoE model trustworthiness with gradient-based methods and introduces a novel approach (ShiftSmooth) for improved model interpretability and robust explainability, aiding motif identification and localization. 
\end{enumerate}

The focus of this work is to show the superiority of the MoE approach in terms of performance, robustness and explainability. The in-distribution binding site motifs to be identified include the ARID3A (AT-Rich Interaction Domain 3A), FOXM1 (Forkhead Box M1), and GATA3 proteins obtained from the JASPAR database \cite{JASPAR}. This database comprises a collection of non-redundant TFBS. ARID3A is a transcription factor involved in cell cycle regulation and embryonic development, playing crucial roles in B lymphocyte development and pluripotency \cite{an2010loss}. FOXM1 is a key regulator of cell proliferation and is often overexpressed in various cancers, making it an important target for cancer research and potential therapeutic interventions \cite{liao2018regulation}. As mentioned earlier, GATA3 is associated with T-cell development and plays a crucial role in regulating cell growth and division in immature red blood cells. Mutations in these genes can lead to various diseases, such as Diamond-Blackfan anemia, a bone marrow failure syndrome, and dyserythropoietic anemia, which results in a deficiency of red blood cells 
\cite{crispino2017gata, kallen2019acquired}.
The study of these transcription factors highlights their importance in understanding genetic disorders and developing targeted therapies.

To demonstrate the robustness of our MoE approach as compared to the individual expert models, we show better generalization on the new OOD TFBS patterns. To rigorously assess the MoE's ability to generalize beyond their training distribution, we randomly select six OOD transcription factors (TFs) that are not part of the original training set. These OOD TFs (BCLAF1, CTCF, POLR2A, RBBP5, SAP30 and STAT3) are from the ENCODE database \cite{ENCODE, encode2011user}. These OOD TFs can be assessed from the ENCODE portal \footnote{https://www.encodeproject.org/}. They represent diverse binding site patterns and genomic contexts that allow us to evaluate the models' performance on truly novel data. This random selection process ensures an unbiased assessment of the models' generalization capabilities and robustness. 

The DNA sequences from the ENCODE database contain genomic sequences that help identify functional elements in the genome. These DNA sequences, known as ChIP-seq elements \cite{mardis2007chip, park2009chip}, are used to map protein-DNA interactions. ChIP-seq samples are prepared by cross-linking a protein of interest with chromatin, shearing the chromatin-protein complex into fragments, and immunoprecipitating the fragments using an antibody. The cross-link is then reversed, and the resulting ChIP samples are sequenced using high-throughput sequencing platforms and mapped to a reference genome. Genomic regions with enriched ChIP reads indicate protein-DNA interactions and represent potential binding sites.

In this study, we also perform a thorough analysis of model trustworthiness using attribution mapping 
\cite{nielsen2023evalattai, ancona2019gradient} 
thereby highlighting the effectiveness of the novel ShiftSmooth technique for interpreting model behavior. Our work offers an efficient, generalizable, and interpretable solution for TFBS prediction.

\begin{figure*}
    \centering
    \includegraphics[width=\textwidth]{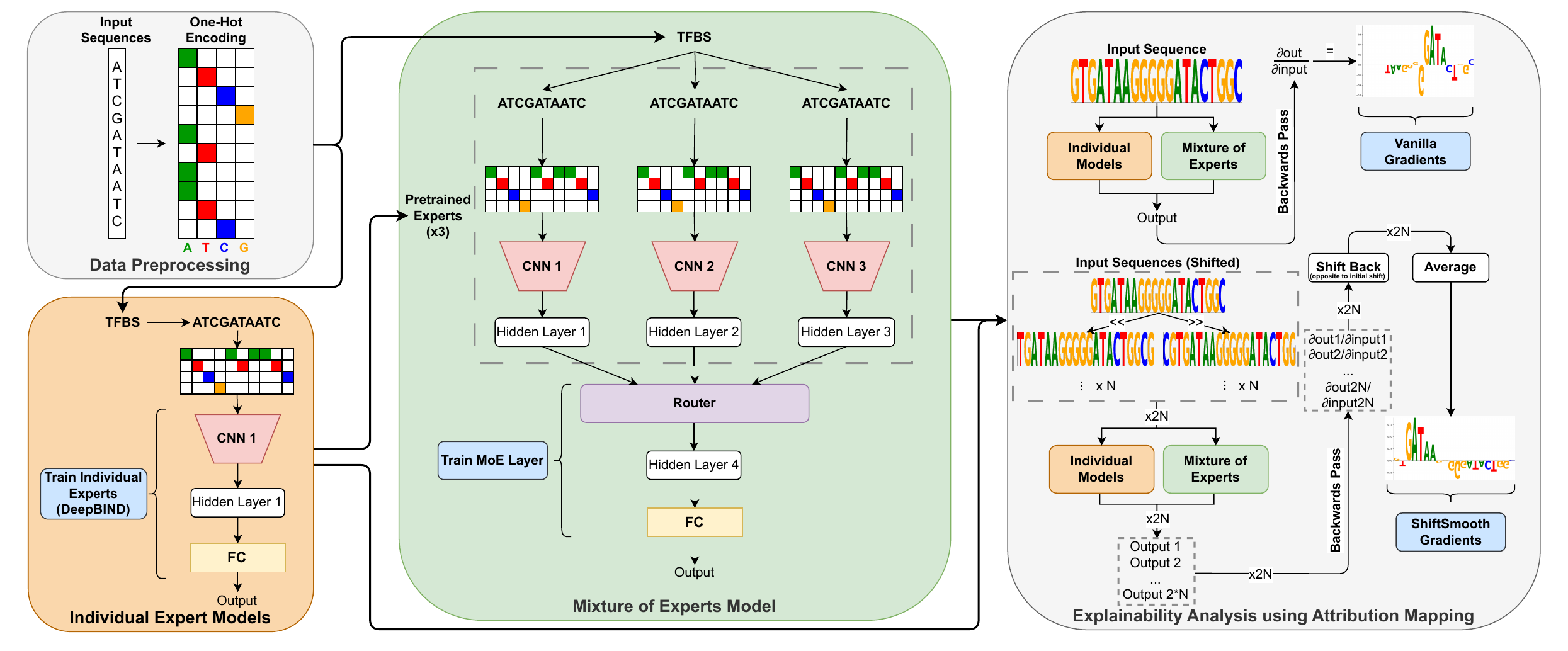}
    \caption{Model training and evaluation process using MoE model and attribution mapping. \textit{Top left}: The data begins as a sequence, then is processed before being input into all models. \textit{Bottom left}: The individual experts are pretrained before being used in the mixture of experts (MoE) model. \textit{Center}: The MoE model is then trained with all expert model weights frozen. \textit{Right}: Finally, all models are evaluated using attribution mapping, including our novel ShiftSmooth method.}
    \label{fig:visualabstract}
\end{figure*}

Figure \ref{fig:visualabstract} provides an overview of the experiments and methods employed in this study. The figure \ref{fig:visualabstract} illustrates the entire process, from data preprocessing to model evaluation. It depicts how the sequence data is processed and fed into individual expert models, which are then integrated into the MoE model. The figure \ref{fig:visualabstract} also illustrates the evaluation process using attribution mapping, including our novel ShiftSmooth method. 

The paper is organized as follows. Section \ref{S:Background} provides background by introducing the model architectures and explainability methods used for analysis and their application to genomic data. In Section \ref{S:Methods}, we present our novel approaches, including the MoE model architecture and the ShiftSmooth explainability method. We also describe the data processing and training procedures for all models. The experimental setup for training, evaluation, and explainability metrics is also detailed in Section \ref{S:Methods}. The results regarding the performance of the models are given in Section \ref{S:Perform}. The explainability results are given in Section \ref{S:Xai}. The conclusions are recorded in Section \ref{S:Conclusion}. 

\section{Background}
\label{S:Background}
This section introduces the model architectures, followed by the specific attribution methods utilized in this work. Related work is also described.

\subsection{Model Architectures}
We consider DeepBIND \cite{deepBIND}, which is a deep learning model designed to predict sequence specificities in genomic data. The reason is that it addresses several challenges associated with learning models for sequencing specificities, such as (1) handling the data formats typically used for genomic data, (2)  processing large amounts of data from high-throughput experiments through parallel implementation, and (3) working effectively across various methods to obtain genomic data. These challenges are often encountered with current technologies that handle genomic data.

The DeepBIND model uses a four-stage process to create a binding score for each input sequence. Note that the binding score quantifies the predicted strength of interaction between a transcription factor and an input DNA/RNA sequence. A higher score indicates stronger binding, while a lower score indicates lesser binding. The four-stage process to create a binding score in the DeepBIND model is briefly discussed here. First, the sequence undergoes convolution to extract local features. Next, a rectification stage applies a non-linear activation function to introduce non-linearity into the model. The third stage involves pooling to reduce the spatial dimensions of the feature maps and retain the most salient features. Finally, the pooled features are passed through a non-linear neural network to produce the binding score. During training, backpropagation is used to update the model's parameters. DeepBIND can be trained on in-vitro data to predict specificities in in-vivo data and vice versa, making it a versatile tool for studying the effects of genetic variants on disease susceptibility.

While DeepBIND has shown good performance in predicting sequence specificities, recent advances in machine learning have introduced more efficient architectures that can achieve similar or better accuracy at a lower computational cost. One such approach is the Mixture of Experts (MoE) layer \cite{sanseviero2023moe}, which was initially proposed decades ago \cite{jacobs1991adaptive} but has only recently been implemented in modern deep learning \cite{eigen2013learning, shazeer2016outrageously}.

Typical MoE models use a ``one-input-to-many-experts" (1:N) setup, where a single input is dynamically routed to a subset of experts, each specializing in different aspects of the input. We utilize the MoE layer in our work. 
However, our model employs a ``many-experts-to-one-output" (N:1) configuration. Each expert processes the entire input independently in this setup, based on the DeepBIND architecture and pre-trained on different transcription factor binding site (TFBS) patterns. Instead of routing inputs, our gating mechanism weights the outputs of all experts to produce a final prediction, allowing simultaneous leverage of each expert's strengths. This N:1 approach offers several advantages. First, it is more flexible, allowing for easy incorporation of new experts without full model retraining. Second, it is more interpretable, as we can analyze which experts contribute most to each prediction. Third, it is computationally efficient, as expert weights remain frozen during MoE training, thereby focusing resources on optimizing the gating network. By employing this N:1 conditional gating approach, our MoE layer (1) significantly improves model efficiency while maintaining accuracy and (2) effectively combines specialized knowledge from multiple experts for better generalization across diverse TFBS patterns.

\subsection{Attribution}

Most state-of-the-art image processing models are considered a black-box, with an internal decision-making process that appears entirely unintelligible to humans. Attribution is used to create visual explanations of which input features are given the most attention by a neural network when making a decision for a particular input. For this reason, they are also commonly referred to as attention maps or feature importance maps. This is a powerful tool since it enables increased user trust in the model. Additionally, attribution can provide key insights into where the model may have gone wrong in cases of incorrect classifications, making it a valuable resource for debugging and developing better models.

\begin{figure}
    \centering
    \includegraphics[width=\columnwidth]{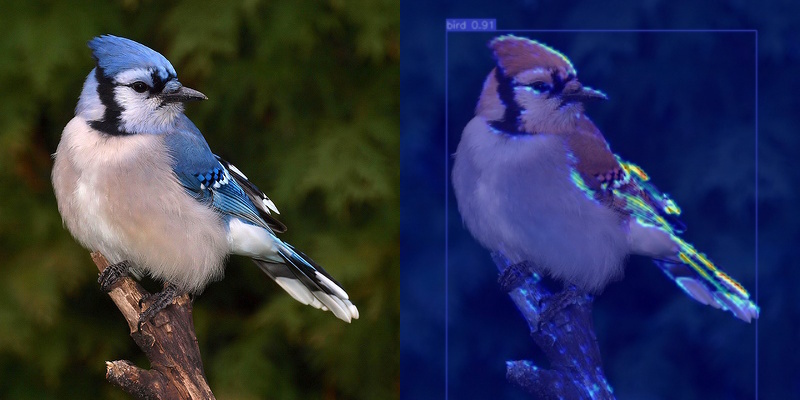}
    \caption{Image input into YOLOv7 \cite{wang2023yolov7} (left) and image overlaid with bounding box prediction and attribution heatmap generated using Grad-CAM \cite{selvaraju2017gradcam} (right). The explanation shows that the model attention is highly focused on the feathers in the image, indicating that the model has learned this as a key feature for the class of bird. Bird image is taken from the COCO dataset \cite{lin2014microsoftcoco}.}
    \label{fig:attribution_example}
\end{figure}

Analysis of deep learning networks using attribution mapping can be beneficial for many tasks, including image classification and object detection. For instance, consider the image of a bird in Fig. \ref{fig:attribution_example}, which is provided as an input to the YOLOv7 \cite{wang2023yolov7} neural network. The model outputs confidence scores and bounding boxes for each class, which can be seen in the right side of Fig. \ref{fig:attribution_example} as predicting bird with 91\% confidence. The popular attribution method known as Grad-CAM \cite{selvaraju2017gradcam} is commonly used to highlight the most important features in the image, which is also shown in Fig. \ref{fig:attribution_example} as a heatmap overlaid onto the input. In this case, the network identified the feathers of the bird to be the most important feature as highlighted in yellow and red. By analyzing these attributions, we can increase our confidence that the model is learning relevant features for this specific image. 

This paper utilizes attribution to explain the comprehensive decision-making process of each model tested thereby enhancing understanding and facilitating the localization of motifs within sequences. Grad-CAM is excellent at highlighting the object of interest in many instances, which is why it is used to depict attribution on an image example. However, it is designed to explain only a specific layer, which makes it difficult to interpret global model knowledge with any single explanation \cite{ali2023explainable}. Hence, the explanation offered by Grad-CAM is limited to fully visualizing the key features learned by the model \cite{nielsen2023evalattai}. In this study, our objective is to explain the decision-making process for the entirety of each model tested, which leads us to exclude this method from our results. Despite its limitations, Grad-CAM retains its utility in applications where visualizing key layers is beneficial. For our testing purposes, we selected two gradient-based attribution methods that visualize the learned features across \textit{all} layers (described in detail later). The first is the Vanilla Gradient. The second is our newly proposed ShiftSmooth Gradient.

Gradient-based attribution methods are used because they are fast and have the advantage of being able to provide detailed and continuous insights into model predictions, and are generally applicable across a wide range of models. Attributional robustness refers to the ability of a method to remain stable and unaffected by minor perturbations or anomalies in the input that do not influence the final decision \cite{nielsen2022robust}. This is important for explainability because it ensures the stability and reliability of the explanations provided by a model. Another key term is plausibility, which describes how convincing an explanation appears to a human observer \cite{nielsen2022robust, nauta2022anecdotal}. As we show in the results, ShiftSmooth enhances attributional robustness and plausibility by taking an average of gradients across multiple modified inputs. In these approaches, the model gradient is used as a starting point to generate explainability maps 
\cite{nielsen2022robust, ancona2019gradient}.
The gradient is acquired by backpropagating the output activation for the class of interest with respect to the input features. For instance, in the image classification example in Fig. \ref{fig:attribution_example}, the output activation would be the bird output logit. The gradient represents how the output activation changes if an input feature is slightly perturbed. 

\subsubsection{Attribution in Genomic Data}
Attribution methods create a map of importance scores attributed to each input feature, providing valuable insight into genomic data and disease risk prediction. The ability to understand what a neural network deems important to a particular decision is essential, especially for mission-critical applications. It has already been established that machine learning has greatly increased the capabilities of medical professionals, but the complexity of models means that interpretability is a major challenge that stands in the way of understanding why certain parameters lead to a particular output. With this in mind, attribution methods are incredibly important in unlocking this ‘black box’, as they have proven to help make the inner workings of a neural network interpretable. 

There are few applications of attribution and explainability for genomic data within the machine learning literature 
\cite{DeepLIFT,dasari2022explainable,clauwaert2021explainability,pope2019explainability}.
Attribution has been proposed as a solution to the problem of discovering motifs that are important in regulatory protein behavior by the authors of DeepLIFT \cite{DeepLIFT}. However, DeepLIFT was not chosen for our testing since (1) it depends on a reference point that is not entirely established \cite{ali2023explainable, sturmfels2020visualizing}, (2) it is not class specific \cite{chefer2021transformer}, and (3) has an open problem with models containing multiple max pooling layers \cite{lundberg2017unified, DeepLIFT}. While we do not use DeepLIFT, we still agree that attribution methods would help in solving this key problem of explainability in the field of computational genomics. As mentioned earlier, we instead use the Vanilla Gradient and ShiftSmooth attribution methods for our explainability assessment.

When explaining genomic data, it is paramount that the explanations are true to what features are actually important to the model. Otherwise, the explanation will not provide credible insight into the model decisions. We will define the criteria of faithfulness for attribution mapping as an assessment of how important the attributed features are to the model itself. For a broad analysis, we refer to the Evaluating Attributions by Adding Incrementally (EvalAttAI) approach \cite{nielsen2023evalattai} which assessed the faithfulness of the most commonly used attribution methods. The EvalAttAI technique \cite{nielsen2023evalattai} finds that the Vanilla Gradient performs the best when assessed for faithfulness. For this reason, the Vanilla Gradient is chosen as one of the attribution methods used. We introduce ShiftSmooth to provide more robust explanations, especially in the context of genome data in which the context window often varies.

\subsubsection{Vanilla Gradient}
Vanilla Gradients (also known as Saliency, sensitivity analysis, or simply the gradient 
\cite{adadi2018peeking, simonyan2014deep, ancona2019gradient, nauta2022anecdotal})
is the first attribution method that we consider in this work. This method is particularly advantageous as it provides a direct explanation of the model without any post-hoc modifications 
\cite{nielsen2022robust, ahmed2023transformers}.
In other words, the gradient directly tells us how the prediction would change if the feature was changed. When compared with other state-of-the-art attribution methods, Vanilla Gradients have demonstrated the ability to generate visualizations that are most representative of the model itself \cite{nielsen2023evalattai}. In many implementations, this method uses the absolute value of the gradient, thereby eliminating sign information. The result is an attribution map that represents the magnitude of pixel importance, irrespective of whether that pixel affects the output activation positively or negatively. While the magnitude is always taken when creating Saliency maps, this is not always the case with Vanilla Gradients. Both methodologies, however, offer valuable insights for elucidating model predictions. In this work, we do not consider the absolute value, thereby allowing us to discern whether each feature impacts the prediction in a positive or negative direction.

The gradient can be defined as an attribution map with ${A}_c^{\text{Gradient}} = \frac{\partial S_c}{\partial \mathbf{x}}$, where $S_c$ is the target class activation score, and $\mathbf{x}$ represents the input features. However, Saliency is a magnitude, so the absolute value of the gradient must be taken. Therefore, we define Saliency maps as ${A}_c^{\text{Saliency}} = \left| \frac{\partial S_c}{\partial \mathbf{x}} \right|$. In this paper, we refer to the absolute value gradient maps as Saliency, and all non-absolute value gradients as Vanilla Gradients.

\subsubsection{ShiftSmooth Gradient}
\label{susub:shiftsmooth}

Machine learning models can often be prone to error when given data that it has not seen before. These differences in data can often come in the form of noise, crop, and other unforeseen variables. To mimic real-world data, it is therefore beneficial to analyze what the attribution map would look like on average, given that noise is random and different each time it is sampled. SmoothGrad \cite{smilkov2017smoothgrad} does this for image processing and is an inspiration for our novel ShiftSmooth Gradient method, which is the second attribution method that we utilize in this work. 

It is important to mention that SmoothGrad, in its original form, cannot be used for explaining TFBS prediction. TFBS prediction differs significantly from image processing, so simply adding some Gaussian noise to the input would not work, as genomic sequences must be given as one of four possible nucleotide types. Furthermore, SmoothGrad also does not account for cropping or shifting of the viewing window of the data, which is especially relevant for genome data since the context window is often shifted to analyze a large sequence in smaller parts. Therefore, we account for this variation by introducing the novel ShiftSmooth method.

ShiftSmooth functions by creating multiple copies of the original data sequence, shifting each one by a different amount, and passing them through the model. It then takes the gradient of each output with respect to each input and shifts the resulting gradient back to the initial starting position (i.e. if the data was shifted to the left twice, then the gradient is shifted to the right twice). Finally, it takes the average of the resulting gradients, creating the ShiftSmooth Gradient. Hence, the obtained attribution maps display to the end user a more robust visualization of what the model would look at on average under realistic circumstances. 



\subsubsection{Robustness}

Robustness can be discussed in the context of machine learning models as a whole, as well as specifically with regards to explainability. A model is said to be robust if its performance shows little decrease when the test data has a different distribution than the data used for training \cite{nielsen2022robust, ancona2019gradient, nielsen2023evalattai}. This can include noisy data, adversarially attacked data, processed or modified data, and data collected in a different context than the training dataset. Models can be made robust through training and through architecture. The former can be achieved by training on a wider distribution of data, often by incorporating noisy or attacked data
\cite{qayyum2020secure, muhammad2022survey, bai2021recent}.
Models can also be robust due to their architecture. For instance, certain probability propagating architectures have been shown to be more robust to noise and adversarial attacks \cite{dera2019extended, dera2021premium}.

In machine learning, a robust model generally achieves the same or similar accuracy when the distribution of the data is modified, while a non-robust model will experience a drop in accuracy. We define robustness in the context of explainability slightly differently than what is done in machine learning. For an explanation to be robust, it should be invariant to small variations that commonly appear in real-world data, such as noise, cropping, and shifting. The ShiftSmooth method takes an average of multiple shifted input gradients to enhance attributional robustness.

Robust models also create more robust and interpretable explanations. Robustly trained models have been shown qualitatively to produce more visually interpretable explanations  
\cite{nielsen2022robust, tsipras2019robustness, nourelahi2022explainable}
that are far less invariant to noise in the data. A similar effect is achieved by SmoothGrad \cite{smilkov2017smoothgrad}, which also produces more robust explanations. Robust training is not always feasible for all applications due to significantly greater computational requirements. Therefore, using robust explainability methods can be a desirable way to create more interpretable visualizations, which aid significantly in the identification of motifs.







\subsection{Related Work}

TFBS has been extensively studied using machine learning, particularly deep learning models, due to their ability to learn complex patterns directly from sequence data. Traditional models such as Position Weight Matrices (PWMs) and supervised learning approaches (e.g., Support Vector Machines (SVMs) and Random Forests) require predefined sequence motifs and heavily rely on manual feature engineering. Although these methods have been foundational, their accuracy drops significantly when dealing with the complexity of gene regulation, as they fail to capture non-linear interactions and do not generalize well across cell types \cite{chen2019interpretable}.

In recent years, convolutional neural networks (CNNs) and RNNs have improved TFBS prediction. DeepBind \cite{deepBIND} and DeepSEA \cite{zhou2015predicting} are well-known CNN-based methods that capture local sequence patterns, offering better performance than traditional approaches by automatically learning from raw genomic sequences. However, one of the main limitations of CNNs is their local receptive field, which hinders their ability to capture long-range dependencies crucial for biological sequence analysis \cite{chen2019interpretable}. DeepSEA addresses this issue partially by incorporating more layers, but this comes at the cost of increased computational complexity and reduced interpretability \cite{lu2024mixture}.

While RNNs, particularly LSTM networks, are better suited for handling sequential dependencies, they tend to be computationally expensive and prone to vanishing gradients over long sequences \cite{chen2019interpretable}. More recently, Transformer-based architectures such as DNA-BERT have been employed to handle long-range interactions using self-attention mechanisms. DNA-BERT leverages pre-trained k-mer embeddings to generalize a broad range of DNA sequences. While this model shows strong generalizability in unsupervised tasks, it often underperforms compared to task-specific models designed for TFBS prediction, particularly in cases where subtle sequence variations are critical \cite{luo2023improving}.

Ensemble models have been increasingly used to combine the strengths of multiple models. The Mixture of Experts (MoE) models have been applied in other areas of bioinformatics, particularly in the prediction of drug-target interactions \cite{lu2024mixture}, where they dynamically assign different experts to different subsets of the input space. The advantage of MoE lies in its ability to leverage multiple specialized models tailored to specific patterns. However, to our knowledge, MoE models have not been applied for the prediction of TFBS. Moreover, most ensemble models tend to focus more on improving performance through aggregation without addressing the challenge of interpretability. This is critical in genomics, where understanding why a model makes a particular prediction is as important as the prediction itself.

Deep learning models for TFBS prediction have traditionally struggled with interpretability. Attribution methods, such as Saliency Maps 
\cite{adadi2018peeking, simonyan2014deep, ancona2019gradient, nauta2022anecdotal},
Integrated Gradients \cite{sundararajan2017axiomatic}, and DeepLIFT \cite{DeepLIFT}, have been proposed to highlight important nucleotides in the input sequence that contribute to the model's decision. However, these methods often fail to provide robust explanations, as they can be highly sensitive to small changes in the input sequence. Furthermore, methods like Grad-CAM \cite{selvaraju2017gradcam}, while useful for interpreting image data, are less applicable to genomic sequences, where subtle shifts in motif location can significantly alter the biological interpretation.

Unlike existing approaches that train a single model per transcription factor (e.g., DeepBind \cite{deepBIND} or DeepSEA \cite{zhou2015predicting}), our model uses a Mixture of Experts (MoE) approach to integrate multiple pre-trained CNN models, each specializing in different TFBS patterns. This setup allows our model to dynamically allocate expert models to specific sequence features based on the input, enabling it to generalize better across different transcription factors. While models like DNA-BERT aim for generalizability through unsupervised learning, our method retains the specificity of task-focused models while extending its utility to out-of-distribution (OOD) transcription factors (TFs)\cite{chen2019interpretable}. Each expert in our MoE model is a pre-trained CNN specialized in recognizing specific TFBS motifs. The MoE framework uses a gating mechanism to weigh the contributions of each expert based on the input sequence, allowing for adaptive specialization depending on the type of TFBS being analyzed. This structure enables the MoE model to dynamically learn which expert is most appropriate for a given transcription factor. This improves performance and flexibility, especially in OOD scenarios where the model encounters unseen TFs \cite{lu2024mixture}.


There are multiple existing attribution methods used for genomic applications which provide a map of importance scores for each nucleotide sequence 
\cite{chandra2024explainable, nauta2022anecdotal, DeepLIFT, ouyang2024developmental}.
These methods include Saliency (i.e. Vanilla Gradients) 
\cite{adadi2018peeking, simonyan2014deep, ancona2019gradient, nauta2022anecdotal},
Integrated Gradients \cite{sundararajan2017axiomatic}, DeepLIFT \cite{DeepLIFT}, DeepSHAP \cite{lundberg2017unified} and silico mutagenesis  
\cite{zhou2015predicting, kelley2016basset}, 
among others. These techniques have been employed to validate whether models have learned features that resemble known motifs across various applications  
\cite{ghanbari2020deep, zhou2015predicting, kelley2016basset}, 
including TFBS motifs 
\cite{ouyang2024developmental, kelley2018sequential, avsec2019deep}. 
Additionally, explainability methods have been utilized for gene characterization and TFBS discovery 
\cite{chandra2024explainable, ouyang2024developmental, amilpur2020edeepssp, novakovsky2023obtaining, novakovsky2023explainn}.

Our method distinguishes itself from existing genomic attribution methods by exclusively using gradient information taken directly from the model to generate explanations, thereby avoiding post-hoc modifications that diverge from the model’s path \cite{nielsen2022robust}. As opposed to post-hoc techniques, our approach involves taking multiple samples of shifted inputs and averaging them to obtain a more robust understanding of what the model considers important in the input. While our approach is somewhat similar to SmoothGrad \cite{smilkov2017smoothgrad}, it can be applied to any task, including those involving genomic sequences as inputs.



\section{Methodology}
\label{S:Methods}

The methodology section details the approaches used in this study. It starts with the model architectures, discusses our proposed MoE model, explains the mathematics behind our novel explainability method (ShiftSmooth) and elaborates on the experimental protocol.

\subsection{Mixture of Experts (MoE) Model}

In this study, we employ a MoE framework to enhance the prediction of DNA-protein binding affinities. The MoE model integrates multiple pre-trained CNN experts, each specialized in capturing distinct features from the input DNA sequences. The architecture consists of two main components: the expert networks and a gating network that dynamically combines the output of each expert.

\subsubsection{Expert Networks}\label{subsub:ExpertNetworks}

Let $N_e$ denote the number of expert models (in our case, $N_e = 3$), each providing embeddings of size $E = 32$. The input DNA sequences are processed by each expert CNN, resulting in a set of embeddings $\{\mathbf{e}_i\}_{i=1}^{N_e}$, where $\mathbf{e}_i \in \mathbb{R}^{B \times E}$ and $B$ is the batch size. Only during the training of the individual experts does each expert model apply a linear transformation followed by a ReLU activation as given by
\begin{equation}
    \mathbf{h}_i = \phi(\mathbf{e}_i \mathbf{W}_{\text{expert}_i} + \mathbf{\beta}),
\end{equation}
where $H = 32$ is the hidden dimension, 
$\mathbf{W}_{\text{expert}_i} \in \mathbb{R}^{E \times H}$ is the weight matrix of the $i$-th expert, 
$\mathbf{\beta} \in \mathbb{R}^{B \times H}$ is a bias matrix in which each row is the bias vector of the $i$-th expert denoted by 
$\mathbf{b}_{\text{expert}_i} \in \mathbb{R}^{1 \times H}$, $\phi(\cdot)$ is the ReLU activation function and
$\mathbf{h}_i \in \mathbb{R}^{B \times H}$ is the output of the $i$-th expert. The bias matrix is expressed as 
\begin{equation}
\label{eq:beta}
   {\mathbf{\beta}} = [\mathbf{b}_{\text{expert}_i}^T ~|~ \mathbf{b}_{\text{expert}_i}^T ~|~ ... ~|~ \mathbf{b}_{\text{expert}_i}^T]^{T}
\end{equation}
Note that the outputs of each expert ($\mathbf{h}_i$ is the output of the $i$th expert) are stacked to form a tensor $H$ given by
\begin{equation}
    \mathbf{H} = [\mathbf{h}_1, \mathbf{h}_2, \mathbf{h}_3] \in \mathbb{R}^{B \times (H \times N_e)} = \mathbb{R}^{B \times (32 \times 3)}.
\end{equation}

It is important to emphasize that for the embedding vectors \( \mathbf{e}_i \) to exist, the hidden representations \( \mathbf{h}_i \) must be learned during the initial training of the expert models. Once trained, we remove the final prediction layer of each expert, effectively using the experts up to their hidden layer as feature extractors in the MoE model.

\subsubsection{Gating Network}

The gating network computes dynamic weights for combining the output of each expert. It takes the concatenated embeddings (denoted by $\mathbf{E}$) from all experts as input to compute the gating scores (denoted by 
$\mathbf{g}$) as expressed by Eqs. \ref{E:Embed} and \ref{E:Gating}.
\begin{equation}
\label{E:Embed}
\mathbf{E} = [\mathbf{e}_1, \mathbf{e}_2, \mathbf{e}_3] \in \mathbb{R}^{B \times (E \times N_e)} = \mathbb{R}^{B \times 96}
\end{equation}
\begin{equation}
\label{E:Gating}
\mathbf{g} = \mathbf{E} \mathbf{W}_{\text{gate}} + \mathbf{\beta}_{\text{gate}} \in \mathbb{R}^{B \times N_e} = \mathbb{R}^{B \times 3}
\end{equation}
In Eq. \ref{E:Gating}, $\mathbf{W}_{\text{gate}} \in \mathbb{R}^{(E \times N_e) \times N_e}$ is a matrix that weights the embeddings of the various experts and $\mathbf{\beta}_{\text{gate}} \in \mathbb{R}^{B \times N_e}$ is a bias matrix in which each row is the bias vector of the weighted embeddings denoted by 
$\mathbf{b}_{gate} \in \mathbb{R}^{1 \times N_e}$.
The bias matrix is expressed as 
\begin{equation}
   \mathbf{\beta_{\text{gate}}} = [\mathbf{b}_{\text{gate}}^T ~|~ \mathbf{b}_{\text{gate}}^T ~|~ ... ~|~ \mathbf{b}_{\text{gate}}^T]^{T}
\end{equation}
The gating weights (denoted by $\boldsymbol{\alpha}$) give the relative emphasis of the output of each expert and are obtained using the softmax function as given by
\begin{equation}
\boldsymbol{\alpha} = \text{softmax}(\mathbf{g}) \in \mathbb{R}^{B \times N_e}.
\end{equation}

\subsubsection{Mixture Output and Prediction}

The final output of the MoE model is a weighted sum of the outputs of each expert. This final mixture output 
$\mathbf{m}$ is calculated as
\begin{equation}
    \mathbf{m} = \sum_{i=1}^{N_e} \boldsymbol{\alpha}_{:,i} \odot \mathbf{h}_i \in \mathbb{R}^{B \times H},
\end{equation}
where $\boldsymbol{\alpha}_{:,i} \in \mathbb{R}^{B \times 1}$ is a column vector that represents the gating weights for the $i$-th expert across the batch, $\mathbf{h}_i \in \mathbb{R}^{B \times H}$ is the output of the $i$-th expert, and $\odot$ denotes element-wise multiplication broadcasted over the batch dimension. This multiplication is performed by taking the $j$th element of 
$\boldsymbol{\alpha}_{:,i}$ and multiplying it by every element in the $j$th row of $\mathbf{h}_i$.  




The mixture output $\mathbf{m}$ is then passed through a classifier that applies a weighting 
($\mathbf{W}_{\text{class}}$) and a bias 
($\mathbf{b}_{\text{class}}$) to get an overall output 
$\mathbf{o}$ as given by
\begin{equation}
    \mathbf{o} = \mathbf{m} \mathbf{W}_{\text{class}} + \mathbf{b}_{\text{class}} \in \mathbb{R}^{B \times 1},
\end{equation}
where $\mathbf{W}_{\text{class}} \in \mathbb{R}^{H \times 1} = \mathbb{R}^{32 \times 1}$ and $\mathbf{b}_{\text{class}} \in \mathbb{R}^{B \times 1}$. 
The final prediction ($\hat{\mathbf{y}}$) is obtained by applying the sigmoid activation function ($\sigma$) to 
$\mathbf{o}$ as given by
\begin{equation}
    \hat{\mathbf{y}} = \sigma(\mathbf{o}) \in \mathbb{R}^{B \times 1}.
\end{equation}

\subsection{Training Procedure}
The training procedure for this methodology is divided into two distinct stages, namely, the training of the individual expert model and the training of the MoE model. The performance of all models is assessed using the ROC curve and its corresponding AUC score.

\subsubsection{Expert Model Training}
The training of the expert models involves a rigorous hyperparameter optimization process. Each CNN expert undergoes optimization using Optuna, an automated hyperparameter optimization framework \cite{optuna}. This process aims to maximize the validation AUC score obtained from the ROC thereby identifying the optimal set of hyperparameters for each expert. The hyperparameter optimization process can be formulated as:
\begin{equation}
    \theta^* = \arg \max_{\theta} \text{AUC}_{\text{val}}(f(\theta))
\end{equation}
where $\theta^*$ represents the optimal hyperparameters, $\theta$ denotes the hyperparameters, $f(\theta)$ represents the CNN model with hyperparameters $\theta$, and $\text{AUC}_{\text{val}}$ is the validation AUC score.

Once the optimal hyperparameters are determined, the CNN experts are trained using these parameters. During this training phase, early stopping is employed to prevent overfitting. This technique monitors the validation performance. If no improvement is observed for a specified number of epochs (set to 5 in this case since the models are individually fine-tuned and optimized for a specific TFBS pattern), the training is halted. This ensures that the model does not continue to train on data that does not enhance its performance, thereby conserving computational resources and preventing overfitting.

\subsubsection{MoE Model Training}
The MoE model is trained by generating embeddings from the pre-trained CNN experts, keeping their weights frozen. The input data is passed through each expert model and the resulting embeddings are concatenated to form a comprehensive representation of the input data.

The MoE model is then trained on these concatenated embeddings using Stochastic Gradient Descent (SGD) with Nesterov momentum and binary cross-entropy loss \cite{sutskever2013importance}. The SGD optimization is defined as
\begin{equation}
    \theta_{t+1} = \theta_t - \eta \nabla L(\theta_t)
\end{equation}
where $\theta_t$ represents the model parameters at iteration $t$, $\eta$ is the learning rate, $L(\theta_t)$ is the loss function, and $\nabla L(\theta_t)$ is the gradient of the loss function with respect to the parameters. The Nesterov momentum is given as
\begin{equation}
    \mathbf{v}_{t+1} = \mu \mathbf{v}_t + \eta \nabla L(\theta_t - \mu \mathbf{v}_t)
\end{equation}
where $\mu$ is the momentum term and $\mathbf{v}_t$ is the velocity vector at iteration $t$.

Early stopping is again applied, but in this case with 10 epochs to ensure that the training is halted if no improvement in validation performance is observed. A patience of 10 epochs allows more flexibility during training, as it adjusts weights across multiple experts and refines the gating mechanism for optimal collaboration.

\subsubsection{Implementation Details}

The MoE model and expert networks are implemented using PyTorch. Training is conducted on GPUs to expedite computation. The MoE model is trained for a maximum of 500 epochs, with early stopping patience set to 10 epochs. The initial learning rate is set to 0.01, and the momentum is set to 0.98. Hyperparameters are fine-tuned using the Optuna framework to optimize the validation AUC.

\subsection{Experimental Setup}

A systematic evaluation framework is implemented alongside a visual explainability analysis to assess the performance of each expert model in comparison to the MoE model. This framework is designed to measure the AUC for each model across multiple test datasets and perform a statistical analysis. This provides a quantitative assessment of the models' generalizability and provides a robust comparison by considering various transcription factor binding site datasets. For each test set, the expert models and the MoE model are evaluated, and their corresponding AUC scores are recorded. 

The AUC metric is chosen for its ability to provide a single measure of the classifier's performance across all classification thresholds, thus capturing the trade-off between true positive and false positive rates. These models are then analyzed using multiple attribution mapping techniques.

\subsubsection{Model Loading and Evaluation}
Each CNN expert model is first evaluated using a test loader that provides batches of ChIP-seq data. The evaluation process involves generating predictions for each batch and calculating the AUC score based on these predictions. This process ensures that each expert model's performance is rigorously assessed using the same evaluation criteria. 

\subsubsection{Bootstrap Method for AUC Estimation}
Bootstrapping \cite{Bishop} is employed to facilitate a statistical analysis and thereby assess the models' generalizability and robustness across various transcription factor binding sites. The approach is to randomly sample, with replacement, from a given test dataset (different from training samples) to generate 30 bootstrapped test datasets. For each bootstrapped test dataset, the AUC score is calculated for all models. This accomplishes 30 trials and generates a distribution of 30 AUC scores for each model. 

\subsubsection{Statistical Analysis}
To determine if the differences in performance between the expert models and the MoE model are statistically significant, an Analysis of Variance (ANOVA) \cite{anova} test is conducted. This statistical comparison of the three expert models and the MoE model will highlight any performance improvements achieved by integrating multiple experts within the MoE framework.

The one-way ANOVA test is based on a null and alternative hypothesis \cite{anova}. 
\begin{enumerate}
\item The null hypothesis states that there is no significant difference between all the four models tested. In this case, the average AUC for all models are essentially equal. The 95\% confidence intervals of all four models overlap. The $p$-value is greater than 0.05.
\item The alternative hypothesis states that there is at least one model with an average AUC that is significantly different than the other three models tested. The 95\% confidence interval of at least one model does not overlap with the other three intervals. The $p$-value is less than 0.05.
\end{enumerate}

\subsection{ShiftSmooth Gradient}
To understand which features matter most to the model, we perform an explainability analysis. To do this, we use the Vanilla Gradient along with our novel gradient-based approach, the ShiftSmooth Gradient, which enhances the robustness of such analysis. The ShiftSmooth method, detailed in Section \ref{susub:shiftsmooth}, is an attribution method that leverages gradients. We use this section to further elaborate on the mathematics behind ShiftSmooth. We define the gradient in Eq. \ref{eq:vanilla_grad} as
\begin{equation}
    A_c (\mathbf{x}) = \frac{\partial S_c (\mathbf{x})}{\partial \mathbf{x}}.
    \label{eq:vanilla_grad}
\end{equation}
Here, $A_c(\mathbf{x})$ represents the model gradient at the input, while $S_c(\mathbf{x})$ denotes the output score for the class of interest $c$, given $\mathbf{x}$ as the model input. The ShiftSmooth method enhances the Vanilla Gradient by computing an average over multiple shifted inputs. The data shifting process is described in Eq. \ref{eq:data_shift} as
\begin{equation}
    \mathbf{x}_n = \mathbf{x} >> n,
    \label{eq:data_shift}
\end{equation}
where $>>$ represents a right bit-shift operation. In Eq. \ref{eq:data_shift}, $x_n$ is the shifted input data, and $n$ indicates the number of places the data is shifted. To limit information loss, elements shifted beyond the final position are reintroduced at the first position (circular shift). The ShiftSmooth Gradient is then calculated by averaging the gradient of all circularly shifted input sequences. To align each individual gradient with the original non-shifted input sequence, the resulting gradient is circularly shifted back $n$ spaces. This is illustrated in Eq. \ref{eq:shiftsmooth_grad} as
\begin{equation}
    \Hat{A_c} (\mathbf{x}) = \frac{1}{2 N + 1} \sum^{N}_{n= -N} A_c (\mathbf{x}_n) << n,
    \label{eq:shiftsmooth_grad}
\end{equation}
where $<<$ indicates a left bit-shift operation. 
The input sequence is shifted backward starting from $n=-N$, and forward up to $n=N$. At $n=0$ no shifting occurs, resulting in the standard non-shifted gradient. The final attribution map ($\Hat{A_c} (\mathbf{x})$), obtained by averaging all these gradients, provides a robust explanation of the model’s focus in the sequence on average, accounting for the variability of the viewing window in real-world data.

\begin{figure}[h]
    \centering
    \includegraphics[width=\columnwidth]{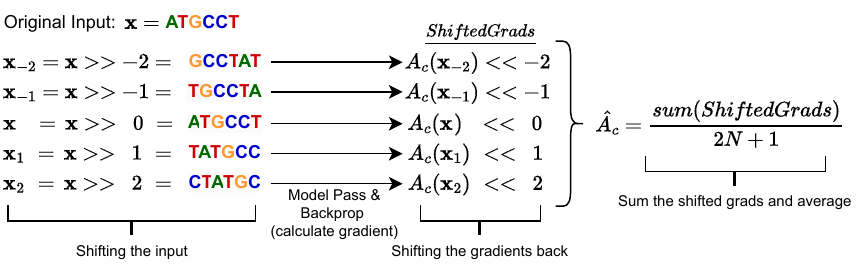}
    \caption{Illustration of the process for generating ShiftSmooth attribution maps $\Hat{A_c}$ for $N = 2$. Starting from the left, the input image is shifted and then passed through the model to compute the gradient. The resulting gradients are subsequently shifted back to reflect the nucleotide locations of the original input. Finally, these shifted gradients are averaged together to produce the resulting ShiftSmooth map $\Hat{A_c}$.}
    \label{fig:shiftsmoothexample}
\end{figure}

The process of generating ShiftSmooth Gradients is detailed in Fig. \ref{fig:shiftsmoothexample}. This figure illustrates four steps: shifting the input (Eq. \ref{eq:data_shift}), calculating the model gradient (Eq. \ref{eq:vanilla_grad}), shifting the gradients back, and taking the average (Eq. \ref{eq:shiftsmooth_grad}). 

First, the input image is shifted as shown in Eq. \ref{eq:data_shift} before the image is passed through the model. Next, the gradient is computed via backpropagation, resulting in $A_c$ as depicted in Eq. \ref{eq:vanilla_grad}. Then, the gradients are shifted to ensure that the attributed features correspond to the positions of the original input data, as shown on the right side of Eq. \ref{eq:shiftsmooth_grad}. Finally, the shifted gradients are averaged to produce the resulting ShiftSmooth Gradient map $\Hat{A_c}$.

\section{Performance Evaluation Results}
\label{S:Perform}

The performance evaluation of the four models is discussed separately for the in-distribution and out-of-distribution (OOD) datasets. The in-distribution datasets refer to the data whose distribution is seen during training for the three expert models. The ARID3A expert model is trained exclusively on ARID3A data. Similarly, the FOXM1 and GATA3 models are trained only on FOXM1 data and GATA3 data respectively. Once trained, the three expert models generate embedding vectors, which are then utilized by the MoE model as described earlier. The MoE model, during its training phase, learns to weight and gate the embedding vectors from all experts but does not have access to raw transcription factor (TF) data. The six OOD datasets (CLAF1, CTCF, POLR2A, RBBP5, SAP30, and STAT3) comprise a distribution not seen during training and give a measure of the robustness of all models.

To rigorously evaluate the significance of the performance differences between the expert models and the MoE model, we conduct a one-way ANOVA \cite{anova} test. This statistical analysis provides a quantitative assessment of the observed differences and helps determine whether they reflect genuine improvements in predictive power or are merely due to chance. 

\subsection{Model Performance on In-Distribution Data}

\begin{figure*}[ht]
    \centering
    \includegraphics[width=1\linewidth]{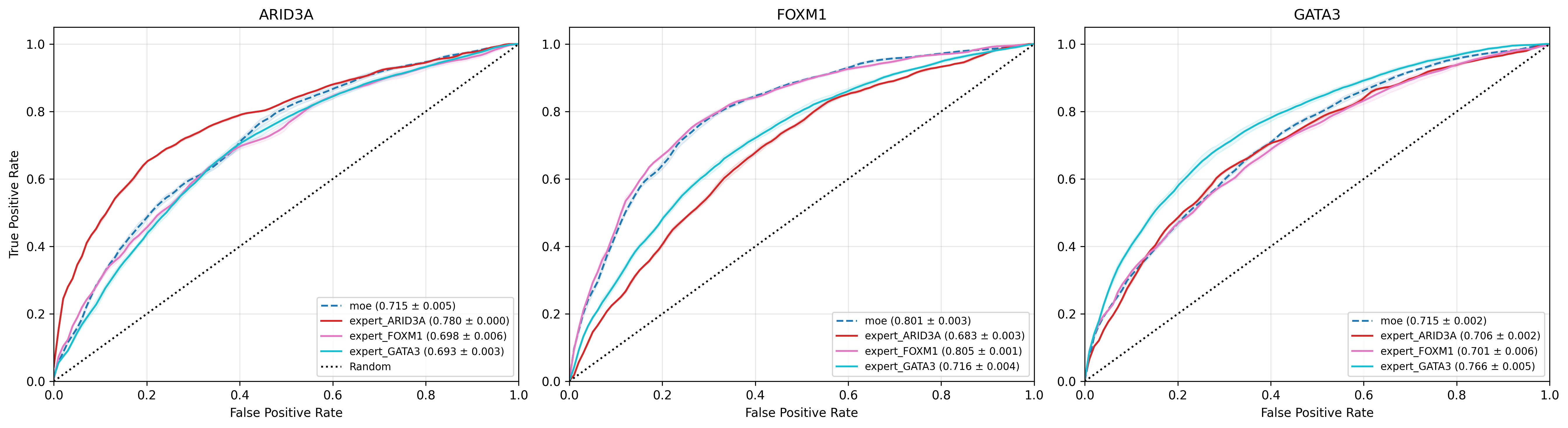}
    \caption{ROC curves for one trial comparing the performance of the ARID3A, FOXM1, GATA3 expert models and the MoE model on in-distribution TFBS prediction.}
    \label{fig:roc}
\end{figure*}

\begin{figure}
    \centering
    \includegraphics[width=1\linewidth]{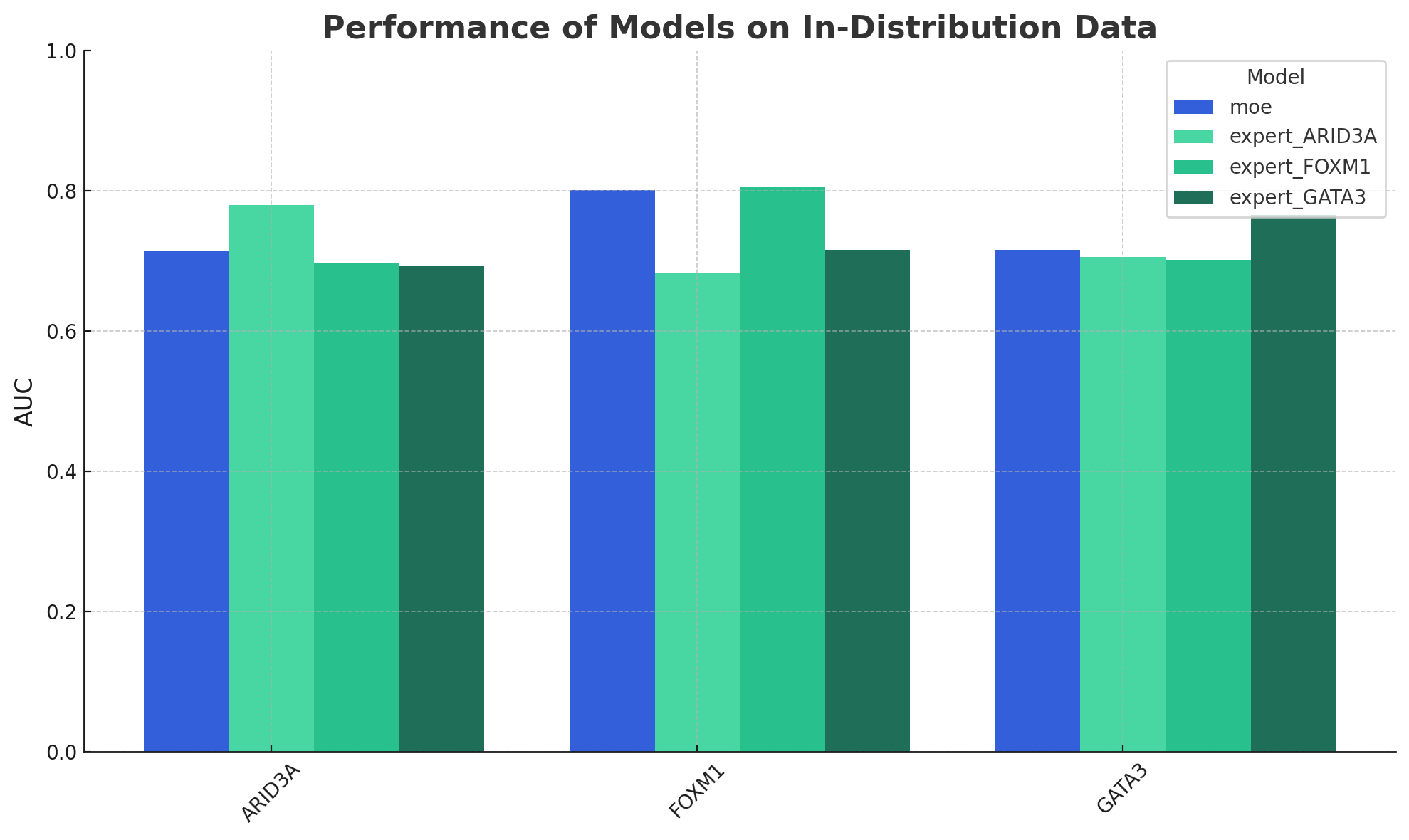}
    \caption{Bar plot showing a comparison of the average model AUC scores taken over 30 trials for each of the in-distribution transcription factor (TF) datasets.}
    \label{fig:in_distribution_bar_plot}
\end{figure}

A comparison of the performance of each model on the three in-distribution datasets (ARID3A, FOXM1, and GATA3) is discussed.
Figure \ref{fig:roc} shows the ROC curves for one trial that compares the (1) ARID3A, FOXM1, and GATA3 motif expert CNN models based on the DeepBIND architecture \cite{deepBIND} and (2) the MoE model. Also shown in Figure \ref{fig:roc} are the mean and standard deviation of the AUC scores (taken over 30 trials) for each model. The standard deviations are low implying consistency among the 30 trials. Figure \ref{fig:in_distribution_bar_plot} depicts a bar plot of the average AUC score for each model again over 30 trials.

\begin{figure}
    \centering
    \includegraphics[width=1\linewidth]{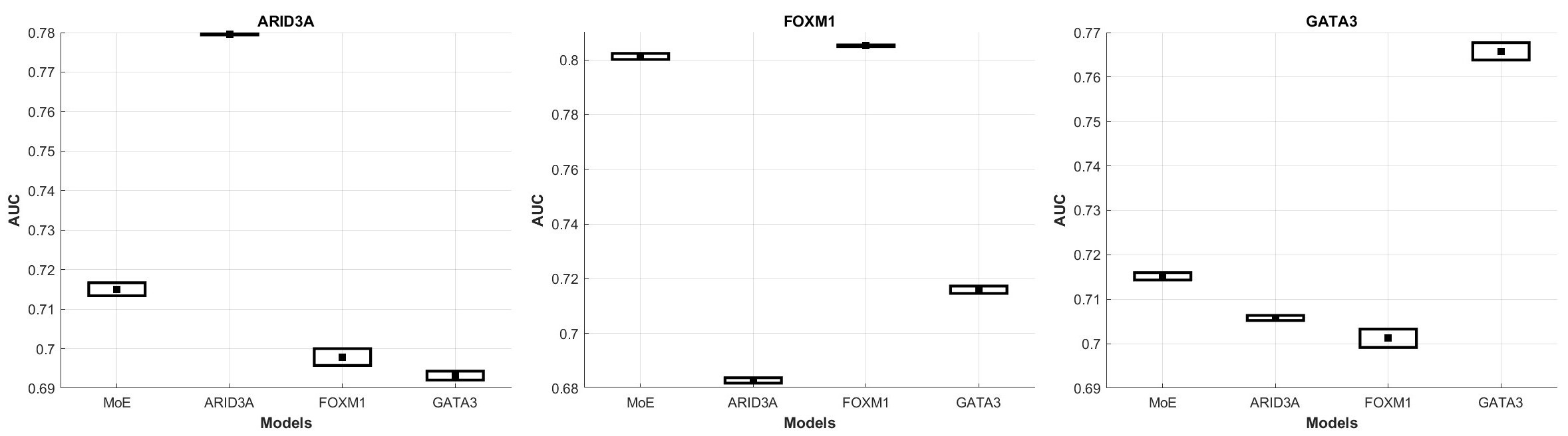}
    \caption{The 95\% confidence intervals (shown as boxes) for the model AUC scores obtained using ANOVA for the ARID3A, FOXM1, and GATA3 datasets.}
    \label{fig:anova_in}
\end{figure}

Each of the expert models shows the best average AUC on their corresponding dataset. The MoE model achieves the second highest average AUC score for all three datasets thereby achieving the best overall solution. 
The ANOVA results support the alternative hypothesis (at least one model is significantly different than the other three) for all three datasets. In fact, (1) there is no overlap among the 95\% confidence intervals of all four models for each of the three datasets and (2) the corresponding $p$-values are extremely small. Figure \ref{fig:anova_in} shows the 95\% confidence intervals of all four models for the ARID3A, FOXM1, and GATA3 datasets. In Figure \ref{fig:anova_in}, some confidence intervals are very thin due to an extremely low standard deviation. Although there is no overlap, some of the intervals are in close proximity. 

It is statistically confirmed that the MoE model offers the best overall solution across the three motifs as it ranks second in all three cases. While we want to emphasize the MoE model's superiority in predicting more complex out-of-distribution motifs as explained in the next section, it is equally important to achieve high performance in predicting simpler in-distribution motifs, as evidenced by the above discussion.

\subsection{Model Performance on Out-of-Distribution Data}

\begin{figure*}[ht]
    \centering
    \includegraphics[width=1\linewidth]{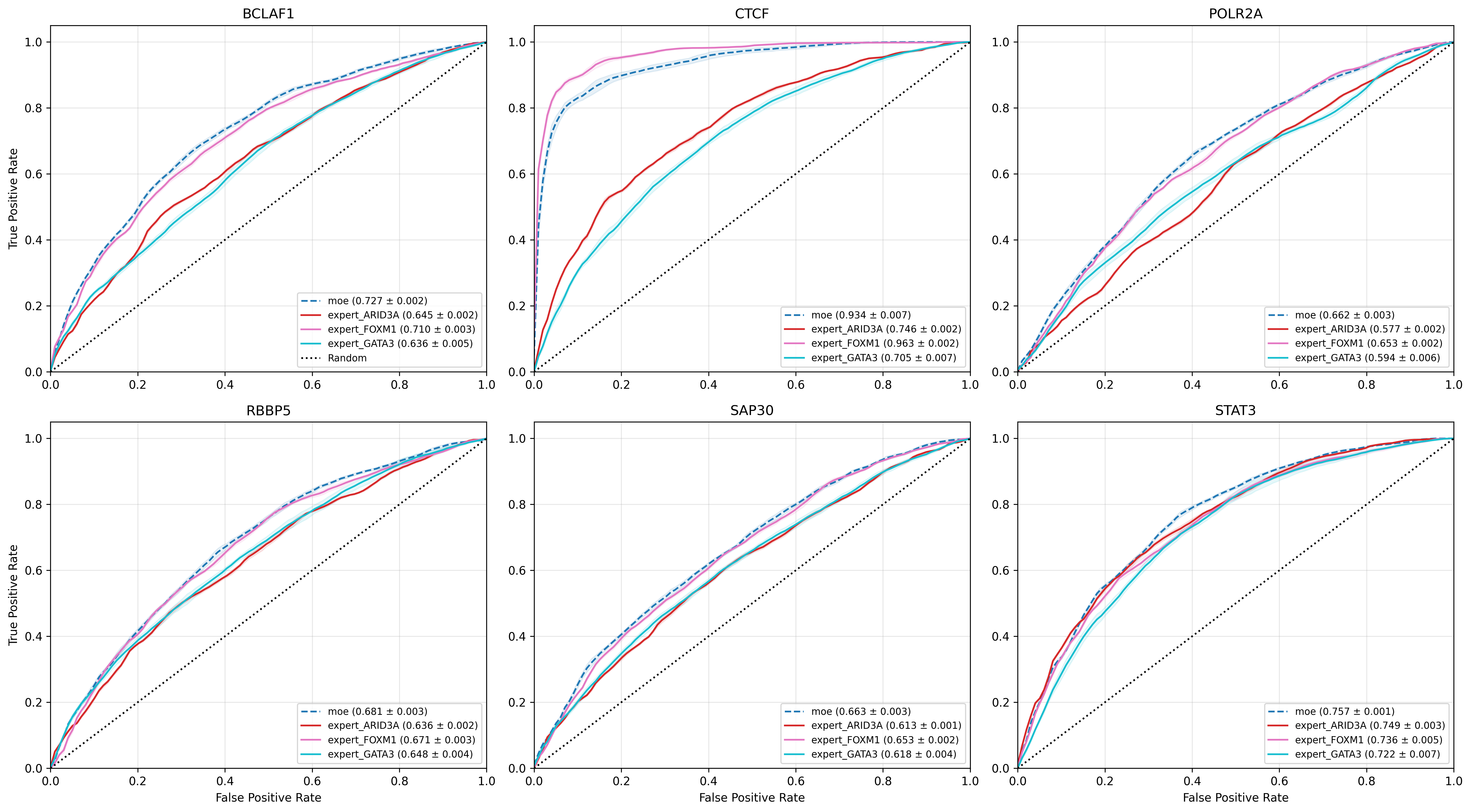}
    \caption{ROC curves comparing the performance of the ARID3A, FOXM1, GATA3 expert models and the MoE model on TFBS prediction across six different OOD transcription factors: BCLAF1, CTCF, POLR2A, RBBP5, SAP30, and STAT3.}
    \label{fig:rocood}
\end{figure*}

Figure \ref{fig:rocood} presents the ROC curves for one trial that compares the ARID3A, FOXM1, and GATA3 motif expert CNN models and the MoE model on TFBS prediction across six different out-of-distribution (OOD) transcription factors (TF). These six datasets represent the binding sites for BCLAF1, CTCF, POLR2A, RBBP5, SAP30, and STAT3. They are randomly selected to assess the models' performance on data that differs from the training distribution. This evaluation is crucial for understanding how well the models generalize to novel TFs and genomic contexts. 

\begin{figure}
    \centering
    \includegraphics[width=1\linewidth]{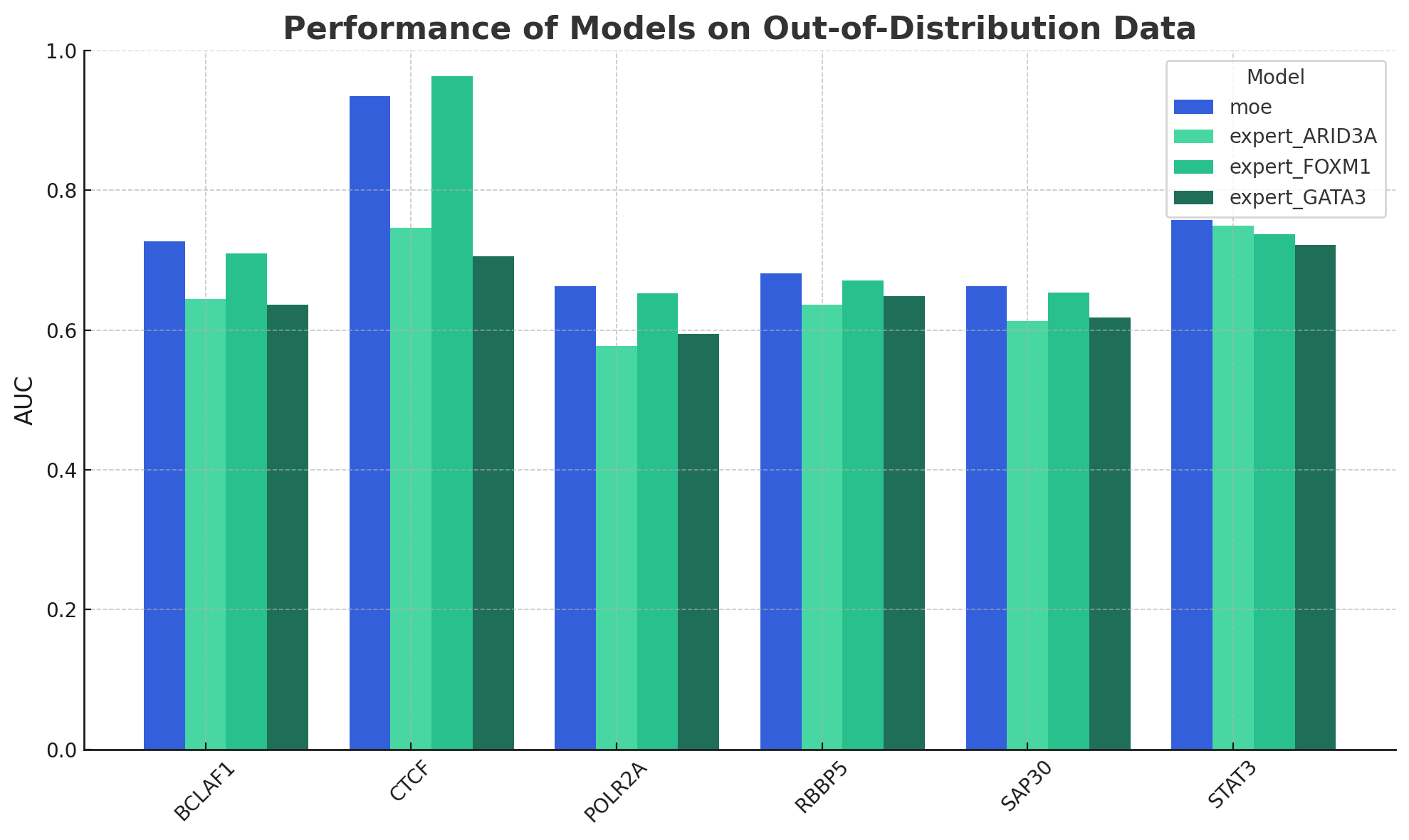}
    \caption{Bar plot showing a comparison of the average model AUC scores taken over 30 trials for each of the out-of-distribution transcription factor (TF) datasets.}
    \label{fig:ood_distribution_bar_plot}
\end{figure}

Figure \ref{fig:ood_distribution_bar_plot} provides a bar plot comparison of model performance on OOD data across the six TF datasets. Specifically, the bars shown in different colors depict the mean AUC scores over 30 trials. As for the in-distribution case, the standard deviations of the AUC scores are extremely low. This again indicates much consistency in the predictions.

\begin{figure}
    \centering
    \includegraphics[width=1\linewidth]{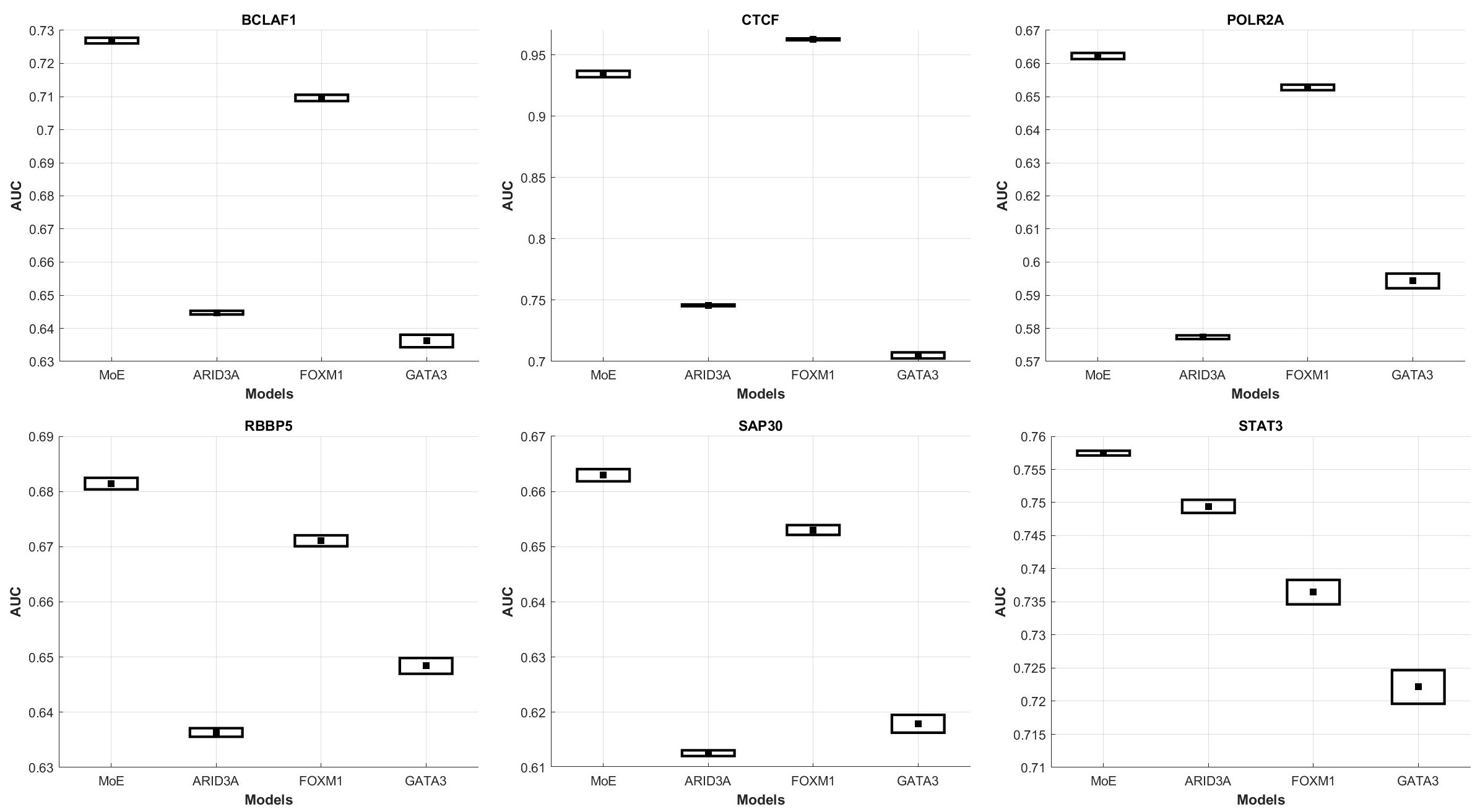}
    \caption{The 95\% confidence intervals (shown as boxes) for the model AUC scores obtained usine ANOVA for the six OOD datasets.}
    \label{fig:anova_out}
\end{figure}

The ANOVA results yield conclusions similar to those in the in-distribution case. Once again, the alternative hypothesis (at least one model is significantly different than the other three) is supported for all six OOD datasets. Also, (1) there is no overlap among the 95\% confidence intervals of all four models for each of the six datasets and (2) the corresponding $p$-values are extremely small. Figure \ref{fig:anova_out} shows the 95\% confidence intervals of all four models for each of the six OOD datasets. 

As illustrated in Figure \ref{fig:anova_out}, the performance of the models varies across the different OOD transcription factors. The MoE is the best for five of the six datasets with statistical significance. It is the second best for the CTCF dataset again with statistical significance. The high performance of the MoE model emphasizes the importance of combining the strengths of various expert models in an ensemble approach. 

The MoE model mitigates the performance fluctuations particularly those exhibited by the ARID3A and the GATA3 expert models thereby demonstrating more stable and consistent performance across all OOD datasets. The FOXM1 model presents the greatest challenge to the MoE, as it achieves the best performance on the CTCF dataset, the second best performance on four datasets, and  third place on the STAT3 dataset. This consistency of the MoE highlights its ability to balance the strengths and weaknesses of individual expert models thereby resulting in more robust overall performance. These findings underscore the value of employing an ensemble approach like the MoE, which leverages the strengths of multiple expert models. By integrating the knowledge and capabilities of these specialized models, the MoE can make informed predictions across a wide range of TFs and genomic contexts, even if it does not always outperform the best individual expert for each specific dataset.

\section{Model Explainability} 
\label{S:Xai}
This section presents a comparative analysis of the explainability results derived from the Vanilla Gradient and ShiftSmooth methods. The study evaluates the attributions generated by these methods when applied to individual expert models and MoE architecture models. By examining the differences in the explanations provided by Vanilla Gradients and ShiftSmooth, we aim to elucidate the strengths and weaknesses of each approach in the context of both model types. For all figures in this section (Figs. \ref{fig:seq_GATA} through \ref{fig:moe_random_ShSm}), the x-axis corresponds to the nucleotide position within the sequence. In the attribution visualizations (Figs. \ref{fig:GATA_VG} to \ref{fig:MoE_ShSm} and Figs. \ref{fig:random_VG} to \ref{fig:moe_random_ShSm}), the y-axis represents the importance scores attributed to each nucleotide.

\subsection{GATA3 Data}

\begin{figure*}[ht]
    \centering
    \includegraphics[width=\textwidth]{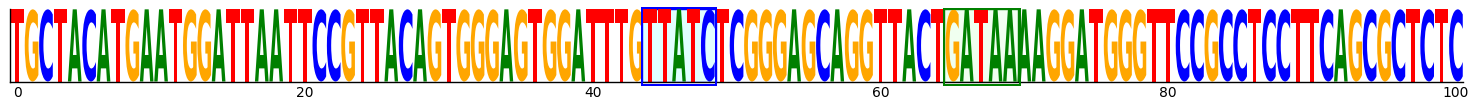}
    \caption{Nucleotide sequence containing GATA3 TFBS. The sequence contains both orientations of the motif including `GATAA' and its inverse, `TTATC.' This sequence is from the ENCODE \cite{ENCODE} database.}
    \label{fig:seq_GATA}
\end{figure*}

To elucidate the differences between the individual expert model and our MoE architecture, we utilize attribution maps to interpret the predictions of both models. Our analysis centers on the input sequence depicted in Fig. \ref{fig:seq_GATA}, which is from the ENCODE database \cite{ENCODE}.

\subsubsection{Expert Model}

\begin{figure*}[ht]
    \centering
    \includegraphics[width=\textwidth]{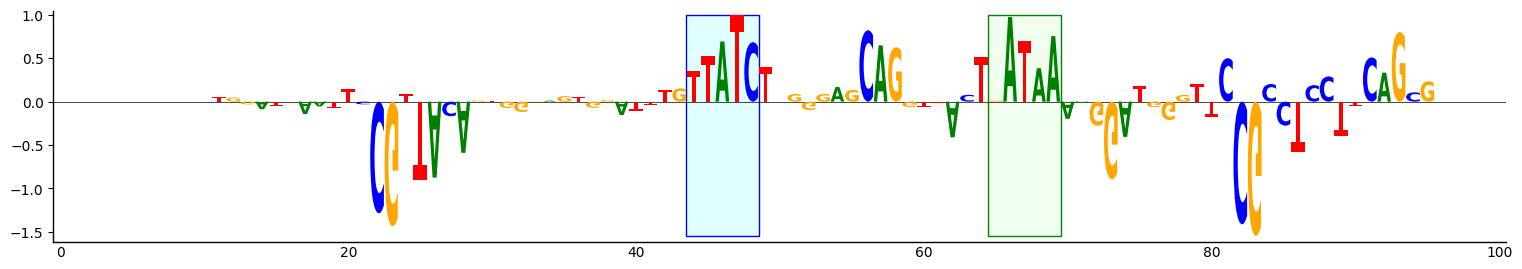}
    \caption{TFBS prediction using the Vanilla Gradient output from the GATA3 trained expert model for the given sequence in Fig. \ref{fig:seq_GATA}.}
    \label{fig:GATA_VG}
\end{figure*}

\begin{figure*}[ht]
    \centering
    \includegraphics[width=\textwidth]{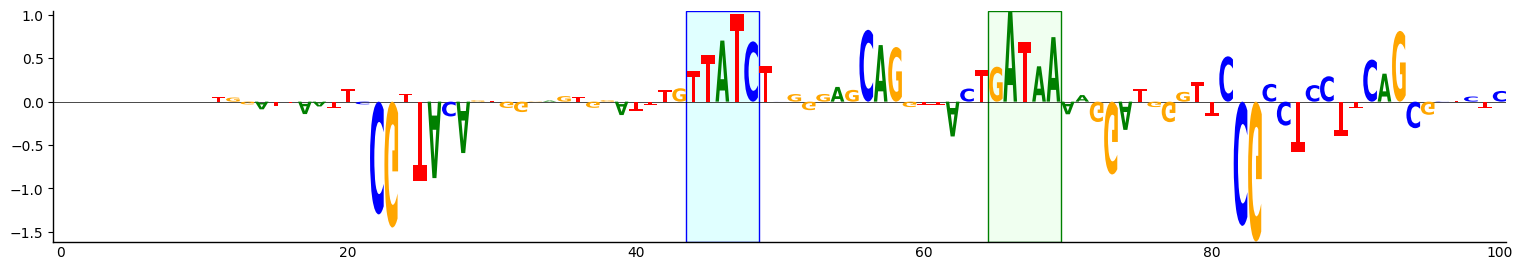}
    \caption{TFBS prediction using the ShiftSmooth Gradient output from the GATA3 trained expert model for the given sequence in Fig. \ref{fig:seq_GATA}.}
    \label{fig:GATA_ShSm}
\end{figure*}

The input sequence is fed into the GATA3-trained expert model, which correctly identifies the presence of the motif. This includes the motif sequence `GATAA' (highlighted in green in Figs. \ref{fig:seq_GATA} to \ref{fig:MoE_ShSm}) and the corresponding inverse sequence, `TTATC' (highlighted in blue in Figs. \ref{fig:seq_GATA} to \ref{fig:MoE_ShSm}). The prediction is visualized using Vanilla Gradients and ShiftSmooth in Figs. \ref{fig:GATA_VG} and \ref{fig:GATA_ShSm}, respectively. 

Figure \ref{fig:GATA_VG} depicts the importance scores attributed to each input by the Vanilla Gradient. The green box shows the scores for the `GATAA' sequence. One can note that the `G' in `GATAA' receives minimal or practically zero importance. The blue box highlights the scores for the `TTATC' sequence. All nucleotides show a positive importance score, with the maximum score attributed to the third `T'. 

Next, we examine the ShiftSmooth method which is visualized in Fig. \ref{fig:GATA_ShSm}. Compared to the Vanilla Gradient, there is essentially no difference for the `TTATC' sequence. This indicates that minor shifts in the input do not affect the attribution for this portion of the input sequence. For the `GATAA' sequence, ShiftSmooth offers a more plausible explanation as indicated by the increased importance attributed to the nucleotide `G' which is essentially absent for the Vanilla Gradient. This also shows promise for the ShiftSmooth method for enhancing the utility of attribution as a tool for motif discovery and localization. 

Given the nature of the ShiftSmooth method, we can show what the model deems important regardless of minor shifts in the data. The viewing window for genomic data is difficult to control, as the start and endpoint of a sequence are somewhat arbitrarily selected for datasets, especially within the margin of a few nucleotides. Therefore, the ShiftSmooth method provides a more attributionally robust explanation of model features, showing how the model will interpret data overall, avoiding the anomalies that arise in the presence of a particular sequence window. This explanation is achieved by aggregating the gradient at multiple context windows and smoothing it out to avoid anomalies. The attributional robustness is exemplified by the ShiftSmooth method assigning importance to the `G,' which is absent in the Vanilla Gradient.

\subsubsection{MoE architecture}

\begin{figure*}[ht]
    \centering
    \includegraphics[width=\textwidth]{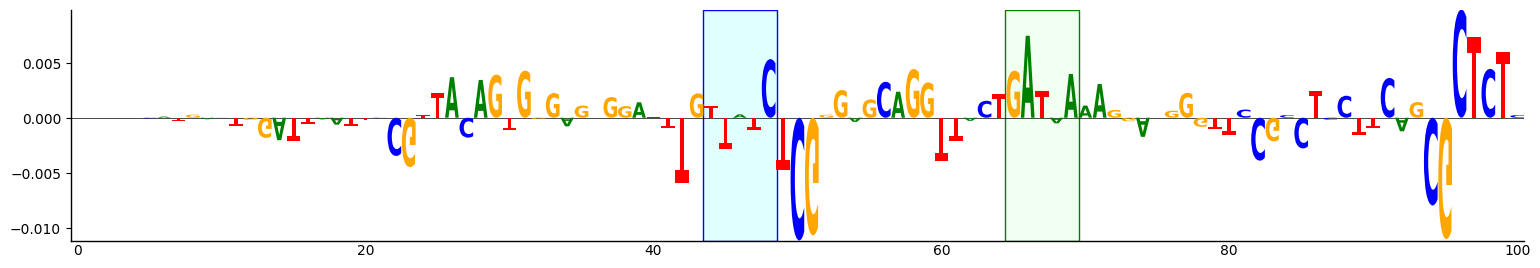}
    \caption{TFBS prediction using the Vanilla Gradient output from the MoE model for the given sequence in Fig. \ref{fig:seq_GATA}.}
    \label{fig:MoE_VG}
\end{figure*}

\begin{figure*}[ht]
    \centering
    \includegraphics[width=\textwidth]{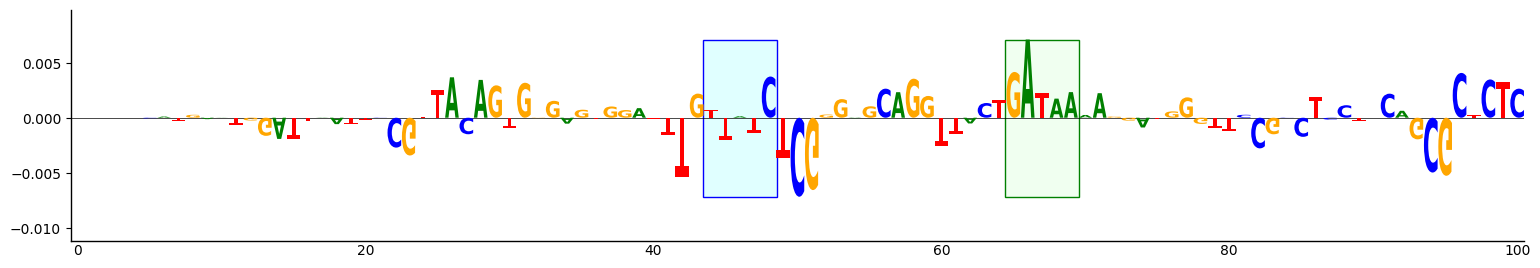}
    \caption{TFBS prediction using the ShiftSmooth Gradient output from the MoE model for the given sequence in Fig. \ref{fig:seq_GATA}.}
    \label{fig:MoE_ShSm}
\end{figure*}

The same sequence shown in Fig. \ref{fig:seq_GATA} is also tested using the MoE model, which correctly identifies the motif. The results are visualized using both attribution methods in Figs. \ref{fig:MoE_VG} and \ref{fig:MoE_ShSm}.
Explaining the individual expert models is straightforward, as each follows a standard CNN architecture. However, explaining the MoE model is slightly more complex due to its three input channels, which remain separate until they converge at the gating layer. 
To explain the complete MoE model, we consider the input sequence as a single tensor sent to all three experts simultaneously, rather than using three identical tensors for each input. By taking the derivative of the MoE output with respect to this single input tensor, we obtain an explanation that represents the complete MoE model. Figure \ref{fig:MoE_VG} presents the explanation of the MoE model obtained using the Vanilla Gradient method. Figure \ref{fig:MoE_ShSm} illustrates the same explanation obtained by the ShiftSmooth technique. 

Figures \ref{fig:MoE_VG} and \ref{fig:MoE_ShSm} reveal that the MoE model is much less focused on any particular motif, aligning with its training to detect all three types of motifs tested. Regarding the `GATAA' motif, the Vanilla Gradient shows the second `A' having a negative attribution. A negative attribution score means that the explanation is less plausible. This is undesirable because it implies that the model is likely looking at less relevant features when making its decision. 
The ShiftSmooth results in Fig. \ref{fig:MoE_ShSm} produced a much more plausible attribution of the importance to the GATA3 motif. There is a positive attribution for the second `A,' which was deemed unimportant by the Vanilla Gradient.
The ShiftSmooth result shows that the negative attribution of the second `A' is an anomaly, as the nucleotide is given a positive attribution when the sequence is shifted slightly. This emphasizes how ShiftSmooth is better for locating key motif features than the Vanilla Gradient. When examining the portion of the sequence highlighted in blue, we see that `TTATC' is given nearly identical attribution scores for both Vanilla Gradient and ShiftSmooth.

Note that for `TTATC,' the attributions for the expert model trained specifically for the `GATAA' sequence (Figs. \ref{fig:GATA_VG} and \ref{fig:GATA_ShSm}) show a high attribution score for all of the nucleotides in `TTATC,' whereas the attributions for the MoE model (Figs. \ref{fig:MoE_VG} and \ref{fig:MoE_ShSm}) show that this is not highly attributed. This shows the specificity of the expert model when compared to the MoE model. The individual expert provides much higher importance to all GATA3 motif features because it is \textit{only} looking for those features.

Overall, we find that ShiftSmooth more effectively attributes the motifs in the sequence as being the most important features for both individual experts and the MoE model. This shows that ShiftSmooth is superior for the localization and identification of motifs.


\subsubsection{Robust Model Training}

While not explored in these results, robust model training creates more interpretable, plausible, and attributionally robust explanations than their non-robust counterparts \cite{tsipras2019robustness, nielsen2022robust}. 
However, robust models require longer training times, and robust pre-trained weights are scarce.
One can achieve some of the benefits of attributionally robust explanations without the need for robust model training.
ShiftSmooth offers these explainability benefits of a robust model without necessitating retraining or additional training time, while still generating  interpretable and plausible results. 
Attributional robustness is particularly significant and beneficial for non-robust models that often attribute noise and irrelevant information as important. 

\subsection{Out-of-Distribution Data}

\begin{figure*}[ht]
    \centering
    \includegraphics[width=\textwidth]{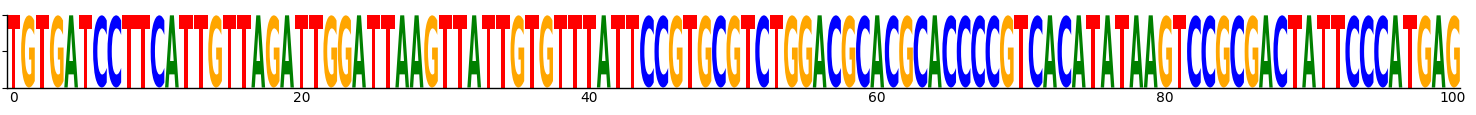}
    \caption{Random nucleotide sequence containing none of the learned motifs.}
    \label{fig:seq_random}
\end{figure*}

To get a better understanding of the key features that each model is focused on, we also analyze the attribution on OOD data. In this case, we generate a random input sequence which does not contain any of the motifs learned by the trained expert models. This sequence can be seen in Fig. \ref{fig:seq_random} which is the input to both the GATA3 trained expert model and the MoE architecture.

\subsubsection{Expert Model}

\begin{figure*}[ht]
    \centering
    \includegraphics[width=\textwidth]{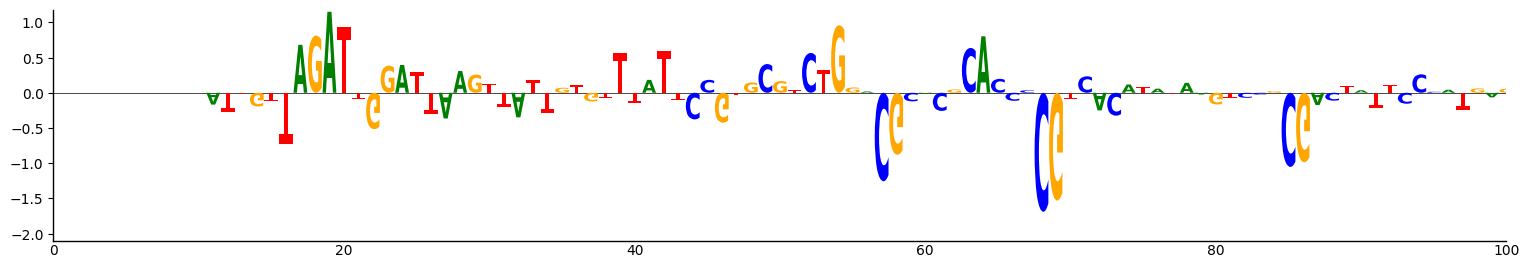}
    \caption{TFBS prediction using the Vanilla Gradient output from the GATA3 trained expert model for the given random sequence in Fig. \ref{fig:seq_random}.}
    \label{fig:random_VG}
\end{figure*}

\begin{figure*}[ht]
    \centering
    \includegraphics[width=\textwidth]{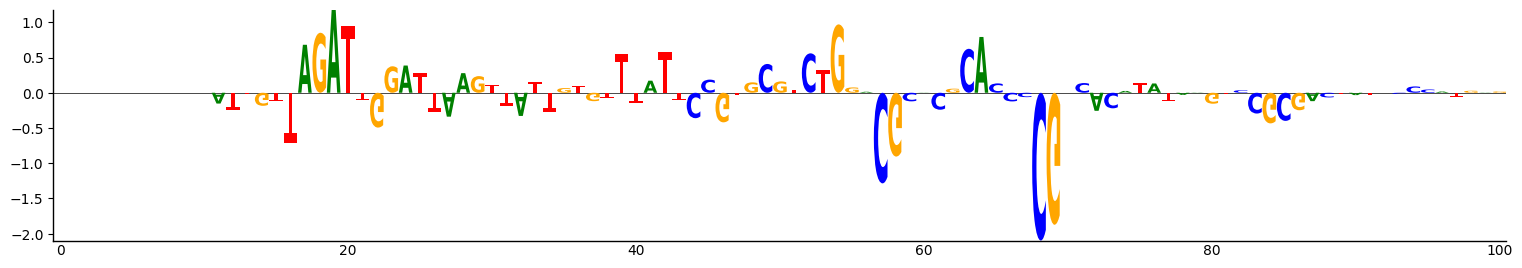}
    \caption{TFBS prediction using the ShiftSmooth Gradient output from the GATA3 trained expert model for the given random sequence in Fig. \ref{fig:seq_random}.}
    \label{fig:random_ShSm}
\end{figure*}

The GATA3 trained expert model correctly predicts that no motif is present. The attribution maps for this prediction are depicted in Figs. \ref{fig:random_VG} and \ref{fig:random_ShSm} using the Vanilla Gradient and ShiftSmooth respectively. Both results show that the model does not deem high importance to any stretches of five or more nucleotides. The model does, however, pick up on pieces of the GATA3 motif, highlighting `GAT' twice between nucleotides 18 and 25 in the sequence. This indicates that the model still considers portions of the motif important, even if the full motif is not present. Note that the Vanilla Gradient and ShiftSmooth attributions present similar results, validating that these same attributions are representative of the model, regardless of small shifts in the selected sequence window. 

\subsubsection{MoE architecture}

\begin{figure*}[ht]
    \centering
    \includegraphics[width=\textwidth]{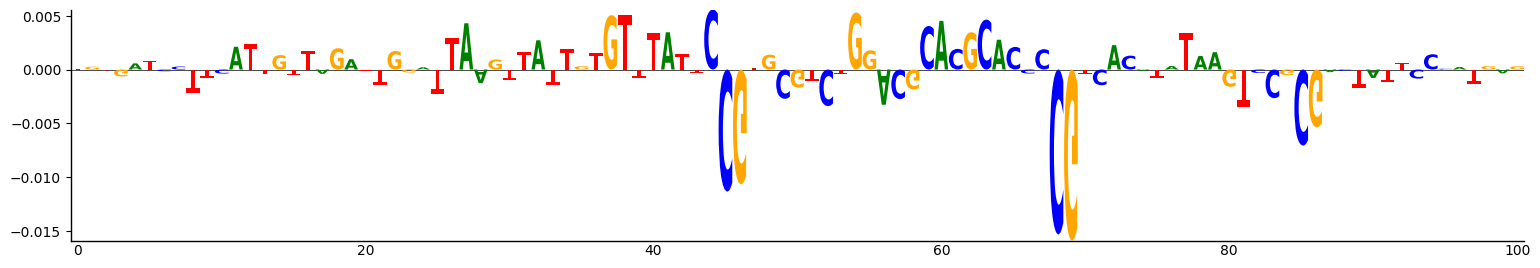}
    \caption{TFBS prediction using the Vanilla Gradient output from the MoE model for the given random sequence in Fig. \ref{fig:seq_random}.}
    \label{fig:moe_random_VG}
\end{figure*}

\begin{figure*}[ht!]
    \centering
    \includegraphics[width=\textwidth]{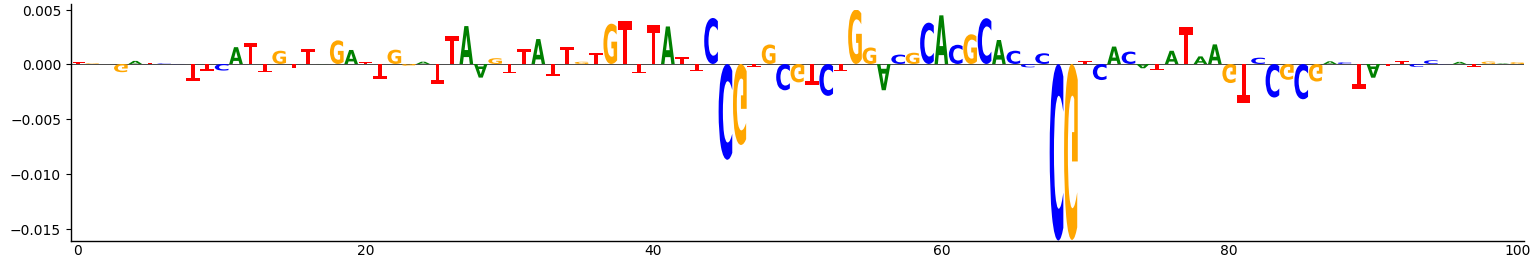}
    \caption{TFBS prediction using the ShiftSmooth Gradient output from the MoE model for the given random sequence in Fig. \ref{fig:seq_random}.}
    \label{fig:moe_random_ShSm}
\end{figure*}

The MoE model also correctly predicts that no motifs are present. The attribution maps corresponding to the Vanilla Gradient and ShiftSmooth methods are shown in Figs. \ref{fig:moe_random_VG} and \ref{fig:moe_random_ShSm}, respectively. Upon examining these visualizations, no stretches of five or more nucleotides corresponding to any of the motif sequences emerge as significantly important. This aligns with expectations, given that the MoE model is looking for multiple motifs, and the sequence lacks any significant features characteristic of any particular learned motif. 
This emphasizes the difference between the specificity of the GATA3 expert and the generality of the MoE model. The GATA3 model is specifically looking for one set of features, whereas the MoE model is looking for multiple. 

We also note that the sequence `CG' consistently has a significant negative attribution for both models and both attribution methods. This is likely because this sequence `CG' is not present in the motif sequences used for training. This was a consistently negatively attributed feature, indicating that this feature likely contributed towards the model \textit{not} predicting the presence of the motif.

\section{Conclusion}
\label{S:Conclusion}

This study introduces a novel Mixture of Experts (MoE) approach for Transcription Factor Binding Site (TFBS) prediction alongside the novel ShiftSmooth explainability method, demonstrating significant advancements in both accuracy and interpretability. Our MoE model, which integrates multiple pre-trained CNN experts, shows superior performance across diverse transcription factor datasets, particularly excelling in generalizing to novel TFBS patterns in OOD scenarios. The statistical analyses provide robust evidence for the significance of the performance differences between the MoE model and individual expert models, highlighting its adaptability to varied genomic contexts. This improved generalization capability is crucial for advancing our understanding of gene regulation across different cell types and organisms.

Another key contribution of this work is the introduction of ShiftSmooth, a novel attribution mapping technique that offers superior attribution for motif discovery and localization compared to the more traditional Vanilla Gradient method. By providing more robust and interpretable insights into the model's decision-making process, ShiftSmooth enhances our ability to identify and localize important TFBS motifs. This advancement in interpretability is critical for building trust in AI-driven genomic analyses and facilitating collaboration between computational biologists and wet lab researchers.

The significance of our work extends beyond computational advancements to real-world applications. In personalized medicine, more accurate TFBS prediction can lead to a better understanding of how genetic variations affect gene regulation, potentially uncovering new drug targets or biomarkers for various diseases. In cancer research, identifying aberrant TFBS patterns could provide insights into tumor development and progression, aiding in discovering new therapeutic targets and developing more effective, personalized treatment strategies. For synthetic biology, accurate TFBS prediction is crucial for designing artificial genetic circuits, with applications ranging from biofuel production to gene therapy. Moreover, the enhanced interpretability offered by ShiftSmooth has significant implications for regulatory compliance in healthcare AI, providing a means to make TFBS prediction models more transparent and auditable.




\bibliographystyle{IEEEtran}

\bibliography{sn-bibliography}{}

\newpage

\vfill

\end{document}